\documentclass[runningheads]{llncs}
\usepackage{graphicx}
\usepackage{tikz}
\usepackage{comment}
\usepackage{amsmath,amssymb}
\usepackage{color}

\usepackage[accsupp]{axessibility}

\definecolor{dark_green}{rgb}{0.0, 0.5, 0.0}
\definecolor{mypurple}{HTML}{7030a0}
\definecolor{myred}{HTML}{ff3b3b}

\usepackage{adjustbox}
\usepackage{multirow}
\usepackage{multicol}
\usepackage{booktabs}
\usepackage{subcaption}
\newlength\savewidth\newcommand\shline{\noalign{\global\savewidth\arrayrulewidth
  \global\arrayrulewidth 1pt}\hline\noalign{\global\arrayrulewidth\savewidth}}

\usepackage[colorlinks=true]{hyperref}
\makeatletter
\newcommand{\phantomlabel}[2]{
    \protected@write\@auxout{}{
        \string\newlabel{#2}{
            {\@currentlabel#1}{\thepage}
            {\@currentlabel#1}{#2}{}
        }
    }
    \hypertarget{#2}{}
}
\makeatother

\begin{document}

\pagestyle{headings}
\mainmatter

\title{
Masked Discrimination for\\Self-Supervised Learning on Point Clouds
}

\def\methodname{MaskPoint}

\titlerunning{Masked Discrimination for Self-Supervised Learning on Point Clouds}

\author{
Haotian Liu \quad
Mu Cai \quad
Yong Jae Lee
}

\authorrunning{Liu et al.}

\institute{
University of Wisconsin--Madison\\
\email{\{lht,mucai,yongjaelee\}@cs.wisc.edu}
}

\maketitle

\begin{abstract}
Masked autoencoding has achieved great success for self-supervised learning in the image and language domains. However, mask based pretraining has yet to show benefits for point cloud understanding, likely due to standard backbones like PointNet being unable to properly handle the training versus testing distribution mismatch introduced by masking during training. In this paper, we bridge this gap by proposing a discriminative mask pretraining Transformer framework, \emph{\methodname{}}, for point clouds. Our key idea is to represent the point cloud as discrete occupancy values (1 if part of the point cloud; 0 if not), and perform simple binary classification between masked object points and sampled noise points as the proxy task. In this way, our approach is robust to the point sampling variance in point clouds, and facilitates learning rich representations. We evaluate our pretrained models across several downstream tasks, including 3D shape classification, segmentation, and real-word object detection, and demonstrate state-of-the-art results while achieving a significant pretraining speedup (e.g., 4.1$\times$ on ScanNet) compared to the prior state-of-the-art Transformer baseline. Code is available at \url{https://github.com/haotian-liu/MaskPoint}.
\end{abstract}

\section{Introduction}

Learning rich feature representations without human supervision, also known as self-supervised learning, has made tremendous strides in recent years. We now have methods in NLP~\cite{Radford2018-GPT,devlin-etal-2019-bert,radford2021learning} and computer vision~\cite{he2020momentum,pmlr-v119-chen20j,mae,chen2021exploring,bao2021beit} that can produce stronger features than those learned on labeled datasets.

In particular, masked autoencoding, whose task is to reconstruct the masked data from the unmasked input (e.g., predicting the masked word in a sentence or masked patch in an image, based on surrounding unmasked context) is the dominant self-supervised learning approach for text understanding~\cite{devlin-etal-2019-bert,joshi2020spanbert,Lan2020ALBERT,yang2019xlnet} and has recently shown great promise in image understanding~\cite{bao2021beit,mae} as well. Curiously, for point cloud data, masked autoencoding has not yet been able to produce a similar level of performance~\cite{occo,yu2021point,yan2022implicit}.  Self-supervised learning would be extremely beneficial for point cloud data, as obtaining high-quality annotations is both hard and expensive, especially for real-world scans.  At the same time, masked autoencoding should also be a good fit for point cloud data, since each point (or group of points) can easily be masked or unmasked. 

We hypothesize that the primary reason why masked autoencoding has thus far not worked well for point cloud data is because standard point cloud backbones are unable to properly handle the distribution mismatch between training and testing data introduced by masking.  Specifically, PointNet type backbones~\cite{qi2017pointnetplusplus,qi2019deep,qi2017pointnet} leverage local aggregation layers that operate over local neighborhoods (e.g., $k$-nearest neighbors) of each point.  The extent of the local neighborhoods can change drastically with the introduction of masking, creating a discrepancy between the distribution of local neighborhoods seen on masked training scenes versus unmasked test scenes. Transformers~\cite{NIPS2017_3f5ee243}, on the other hand, can perform self-attention (a form of aggregation) on either all or selective portions of the input data.  This means that it has the ability to only process the unmasked portions of the scene in the training data, without being impacted by the masked portions.  This property suggests that Transformers could be an ideal backbone choice for self-supervised masked autoencoding for point clouds. 

\begin{figure}[t]
    \centering
    \centering
    \includegraphics[width=.7\textwidth]{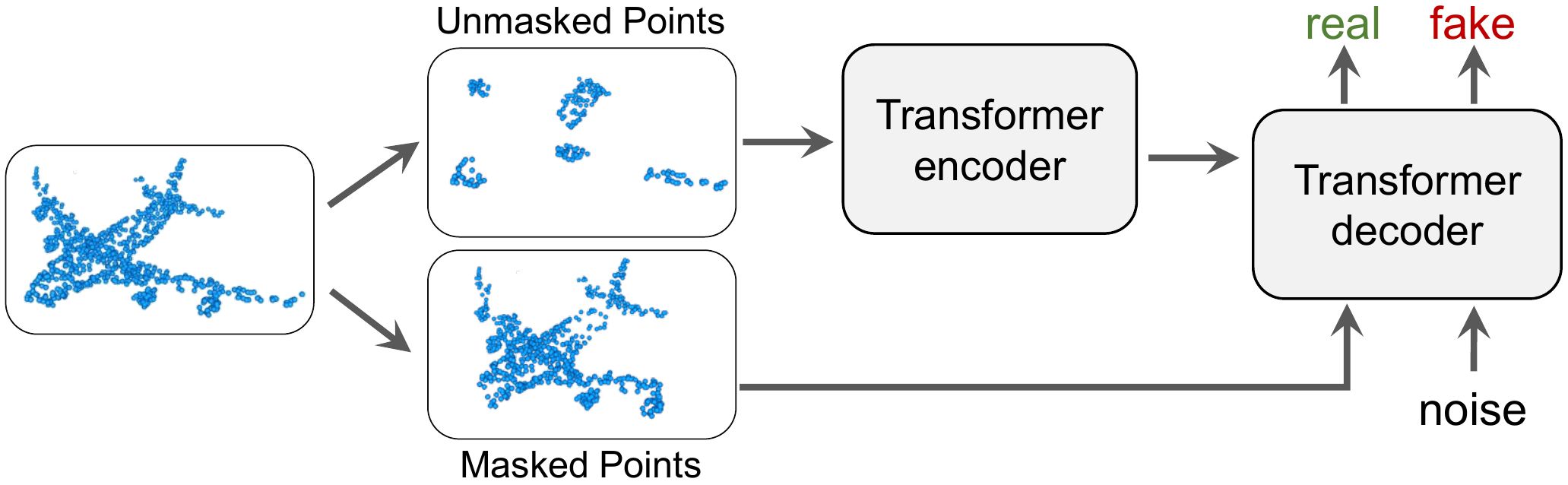}
    \caption{\textbf{Main Idea.}  We randomly partition the point cloud into masked and unmasked sets.  We only feed the visible portion of the point cloud into the encoder.  Then, a set of real query points are sampled from the masked points, and a set of fake query points are randomly sampled from 3D space.  We train the decoder so that it distinguishes between the real and fake points. After pre-training, we discard the decoder and use the encoder for downstream tasks.}
    \label{fig:main_idea}
\end{figure}

For image understanding, the state-of-the-art masked autoencoding Transformer approach MAE~\cite{mae} masks out a large random subset of image patches, applies the Transformer encoder to the unmasked patches, and trains a small Transformer decoder that takes in the positional encodings of the masked patches to reconstruct their original pixel values.  However, this approach cannot be directly applied to point cloud data, because the raw representation of each 3D point is its spatial xyz location. Thus, training the decoder to predict the xyz coordinates of a masked point would be trivial, since its positional encoding would leak the correct answer. In this case, the network would simply take a shortcut and not learn meaningful features.

To address this, we propose a simple binary point classification objective as a new pretext task for point cloud masked autoencoding.  We first group points into local neighborhoods, and then mask out a large random subset of those groups.  The Transformer encoder takes in the unmasked point groups, and encodes each group through self-attention with the other groups. The Transformer decoder takes in a set of real and fake query points, where the real queries are sampled from the \emph{masked} points, while the fake queries are randomly sampled from the full 3D space.  We then perform cross attention between the decoder queries and encoder outputs.  Finally, we apply a binary classification head to the decoder's outputs and require it to distinguish between the real and fake queries.  We find this simple design to be effective, as it creates a difficult and meaningful pretext task that requires the network to deduce the 3D shape of the object from only a small amount of visible point groups.

Among prior self-supervised point cloud approaches, Point-BERT~\cite{yu2021point} is the most related.  It trains a discrete Variational AutoEncoder (dVAE)~\cite{dvae} to encode the input point cloud into discrete point token representations, and performs BERT-style pretraining over them.
To aid training, it uses point patch mixing augmentation, together with an auxiliary MoCo~\cite{he2020momentum} loss.
However, the dependency on a pretrained dVAE together with other auxiliary techniques, creates a significant computational overhead in pretraining -- our experiments show that its pre-training is significantly slower (e.g., $4.1\times$ on ScanNet~\cite{dai2017scannet}) than ours even without taking into account the training time of the dVAE module.
The large speedup is also due to our design of the Transformer encoder processing only the unmasked points.

In sum, our main contributions are: (1) A novel masked point classification Transformer, \emph{\methodname{}}, for self-supervised learning on point clouds. (2) Our approach is simple and effective, achieving state-of-the-art performance on a variety of downstream tasks, including object classification on ModelNet40~\cite{wu20153d} / ScanObjectNN~\cite{uy2019revisiting}, part segmentation on ShapeNetPart~\cite{yi2016scalable}, object detection on ScanNet~\cite{dai2017scannet}, and few-shot object classification on ModelNet40~\cite{wu20153d}. (3) Notably, for the first time, we show that a standard Transformer architecture can outperform sophisticatedly designed point cloud backbones.
\section{Related Work}

\subsubsection{Transformers.} 
Transformers were first proposed to model long-term dependencies in NLP tasks~\cite{NIPS2017_3f5ee243}, and have achieved great success~\cite{NIPS2017_3f5ee243,devlin-etal-2019-bert,Radford2018-GPT}. More recently, they have also shown promising performance on various image and video understanding tasks \cite{dosovitskiy2020vit,tolstikhin2021mixer,steiner2021augreg,carion2020end,max_deeplab_2021,jiang2021transgan,radford2021learning}. There have also been attempts to adopt transformers to 3D point cloud data. PCT~\cite{guo2021pct} and Point Transformer~\cite{zhao2021point} propose new attention mechanisms for point cloud feature aggregation. 3DETR~\cite{misra2021-3detr} uses Transfomer blocks and the parallel decoding strategy from DETR~\cite{carion2020end} for 3D object detection.  However, it is still hard to get promising performance using the standard Transformer. For example, in 3D object detection, there is a large performance gap between 3DETR~\cite{misra2021-3detr} and state-of-the-art point based~\cite{zhang2020h3dnet} and convolution based~\cite{rukhovich2021fcaf3d} methods. In this paper, we show that self-supervised learning using a novel masked point classification objective can aid a standard Transformer to learn rich point cloud feature representations, and in turn largely improve its performance on various downstream tasks.

\vspace{-10pt}
\subsubsection{Self-supervised Learning.}
Self-supervised learning  (SSL) aims to learn meaningful representations from the data itself, to better serve downstream tasks. Traditional methods typically rely on pretext tasks, such as image rotation
prediction~\cite{gidaris2018unsupervised}, image colorization~\cite{zhang2016colorful}, and solving jigsaw puzzles~\cite{noroozi2016unsupervised}.  Recent methods based on contrastive learning (e.g., MoCo~\cite{he2020momentum}, SimCLR~\cite{pmlr-v119-chen20j}, SimSiam~\cite{chen2021exploring}) have achieved great success in the image domain, sometimes producing even better downstream performance compared to  supervised pretraining on ImageNet~\cite{deng2009imagenet}.

Self-supervised learning has also begun to be explored for point cloud data. Pretext methods include deformation reconstruction~\cite{achituve2021self}, geometric structure prediction~\cite{thabet2020self}, and orientation estimation~\cite{poursaeed2020self}. Contrastive learning approaches include  PointContrast~\cite{xie2020pointcontrast}, which learns corresponding points from different camera views, and DepthContrast~\cite{Zhang_2021_ICCV}, which learns representations by comparing transformations of a 3D point cloud/voxel.  OcCo~\cite{occo} learns an autoencoder to reconstruct the scene from the occluded input. However, due to the sampling variance of the underlying 3D shapes, explicitly reconstructing the original point cloud will inevitably capture such variance. In this paper, we explore a simple but effective discriminative classification pretext task to learn representations that are robust to the sampling variance.

\vspace{-10pt}
\subsubsection{Mask based Pretraining.}
Masking out content has been used in various ways to improve model robustness including as a regularizer~\cite{dropout,dropblock}, data augmentation~\cite{devries2017improved,singh2017hide,randomerasing}, and self-supervised learning~\cite{chen2022context,devlin-etal-2019-bert,mae}.  For self-supervised learning, the key idea is to train the model to predict the masked content based on its surrounding context.  The most successful approaches are built upon the Transformer~\cite{NIPS2017_3f5ee243}, due in part to its token-based representation and ability to model long-range dependencies.  

In masked language modeling, BERT~\cite{devlin-etal-2019-bert} and its variants~\cite{joshi2020spanbert,Lan2020ALBERT} achieve state-of-the-art performance across nearly all NLP downstream tasks by predicting masked tokens during pretraining.  Masked image modeling works~\cite{bao2021beit,mae} adopt a similar idea 
for image pretraining. BEiT~\cite{bao2021beit} maps image patches into discrete tokens, then masks a small portion of patches, and feeds the remaining visible patches into the Transformer to reconstruct the tokens of the masked patches. Instead of reconstructing tokens, the recent Masked AutoEncoder (MAE)~\cite{mae} reconstructs the masked patches at the pixel level, and with a much higher mask ratio of $\geq 70\%$. Following works try to predict higher-level visual features such as HoG~\cite{wei2021masked}, or improve the representation capability of the encoder by aligning the feature from both visible patches and masked patches~\cite{chen2022context}. To our knowledge, the only self-supervised mask modeling Transformer approach for point clouds is Point-BERT~\cite{yu2021point}, which adopts a similar idea as BEiT~\cite{bao2021beit}. However, to obtain satisfactory performance, it requires a pretrained dVAE and other auxiliary techniques (e.g., a momentum encoder~\cite{he2020momentum}), which slow down training.  Our masked point discrimination approach largely accelerates training (\textbf{$4.1\times$} faster than Point-BERT) while achieving state-of-the-art performance for various downstream tasks.

Finally, some concurrent works~\cite{pang2022masked} also explore adopting the masked autoencoding pretraining objective on point clouds. Our key novelty over these contemporary approaches is the discriminative point classification objective which helps to address the sampling variance issue in point clouds.

\section{Approach}

The goal is to learn semantic feature representations without human supervision that can perform well on downstream point cloud recognition tasks.
We motivate our self-supervised learning design with a qualitative example.  Fig.~\ref{fig:main_idea} ``Unmasked Points'' shows a point cloud with a large portion (90\%) of its points masked out.  Still, based on our prior semantic understanding of the world, we as humans are able to say a number of things about it: (1) it might be an airplane; (2) if so, it should consist of a head, body, tail, and wings; and even (3) roughly where these parts should be present.  In other words, because we already know what airplanes are, we can recover the missing information from the small visible subset of the point cloud.  In a similar way, training a model to recover information about the masked portion of the point cloud given the visible portion could force the model to learn object semantics.

However, even as humans, it can be difficult or impossible to precisely reconstruct all missing points, since there are several ambiguous factors; e.g., the precise thickness of the wings or the precise length of the airplane. If we are instead given a sampled 3D point in space, and are asked to answer whether it likely belongs to the object or not, we would be more confident in our answer.  This discriminative point classification task is much less ambiguous than the reconstruction task, yet still requires a deep understanding of object semantics in order to deduce the masked points from the small number of visible points.

\subsection{Masked Point Discrimination}\label{sec:dmpm}

Our approach works as follows.  We randomly partition each input point cloud $\mathcal{P} \in \mathbb{R}^{N \times 3}$ into two groups: masked $\mathcal{M}$ and unmasked $\mathcal{U}$.  We use the Transformer encoder to model the correlation between the sparsely-distributed unmasked tokens $\mathcal{U}$ via self-attention.
Ideally, the resulting encoded latent representation tokens $\mathcal{L}$ should not only model the relationship between the unmasked points $\mathcal{U}$, but also recover the latent distribution of masked points $\mathcal{M}$, so as to perform well on the pretraining task.
We next sample a set of \emph{real} query points $\mathbf{Q}_{real}$ and a set of \emph{fake} query points $\mathbf{Q}_{fake}$.  The real query points are sampled from the masked point set $\mathcal{M}$, while the fake query points are randomly sampled from the full 3D space.
We then perform cross attention between each decoder query $\textbf{q} \in \{\mathbf{Q}_{real}, \mathbf{Q}_{fake}\}$ and the  encoder outputs, $\mathtt{CA}(\textbf{q}, \mathcal{L})$, to model the relationship between the masked query point and the unmasked points.
Finally, we apply a binary classification head to the decoder's outputs and require it to distinguish between the real and fake queries.  

We show in our experiments that our approach is both simple and effective, as it creates a pretext task that is difficult and meaningful enough for the model to learn rich semantic point cloud representations.

\vspace{-10pt}
\subsubsection{Discarding Ambiguous Points.}
\label{AmbiguousPts}
Since we sample fake query points uniformly at random over the entire space, there will be some points that fall close to the object's surface. Such points can cause training difficulties since their target label is `fake' even though they are on the object.  In preliminary experiments, we find that such ambiguous points can lead to vanishing gradients in the early stages of training.  Thus, to stabilize training, we simply remove all fake points $\mathbf{\hat{p}} \in \mathbf{Q}_{fake}$ whose euclidean distance is less than $\gamma$ to any object (masked or unmasked) point $\mathbf{p}_i \in \mathcal{P}$: $\min_i ||\mathbf{\hat{p}} - \mathbf{p}_i||_2 < \gamma$.
To address the size variance of the input point cloud, $\gamma$ is dynamically selected per point cloud $\mathcal{P}$: $\hat{\mathcal{P}} = \operatorname{FPS}(\mathcal{P}), \gamma = \min_{j \neq i} ||\hat{\mathcal{P}}_i - \hat{\mathcal{P}}_j||_2$.

\vspace{-10pt}
\subsubsection{3D Point Patchification.}
\label{Point Patchify}
Feeding every single point into the Transformer encoder can yield an unacceptable cost due to the quadratic complexity of self-attention operators.  Following \cite{yu2021point,dosovitskiy2020vit}, we adopt a patch embedding strategy that converts input point clouds into 3D point patches.

Given the input point cloud $\mathcal{P} \in \mathbb{R}^{N \times 3}$, $S$ points $\{\mathbf{p}_i\}_{i=1}^S$ are sampled as patch centers using farthest point sampling~\cite{qi2017pointnetplusplus}.  We then gather the $k$ nearest neighbors for each patch center to generate a set of 3D point patches $\{\mathbf{g}_i\}_{i=1}^{S}$.  A PointNet~\cite{qi2017pointnet} is then applied to encode each 3D point patch $\mathbf{g}_i \in \mathbb{R}^{k \times 3}$ to a feature embedding $\mathbf{f}_i \in \mathbb{R}^d$.  In this way, we obtain $S$ tokens and their corresponding features $\{\mathbf{f}_i\}_{i=1}^S$ and center coordinates $\{\mathbf{p}_i\}_{i=1}^S$.

\begin{figure}[t]
    \centering
    \includegraphics[width=.8\textwidth]{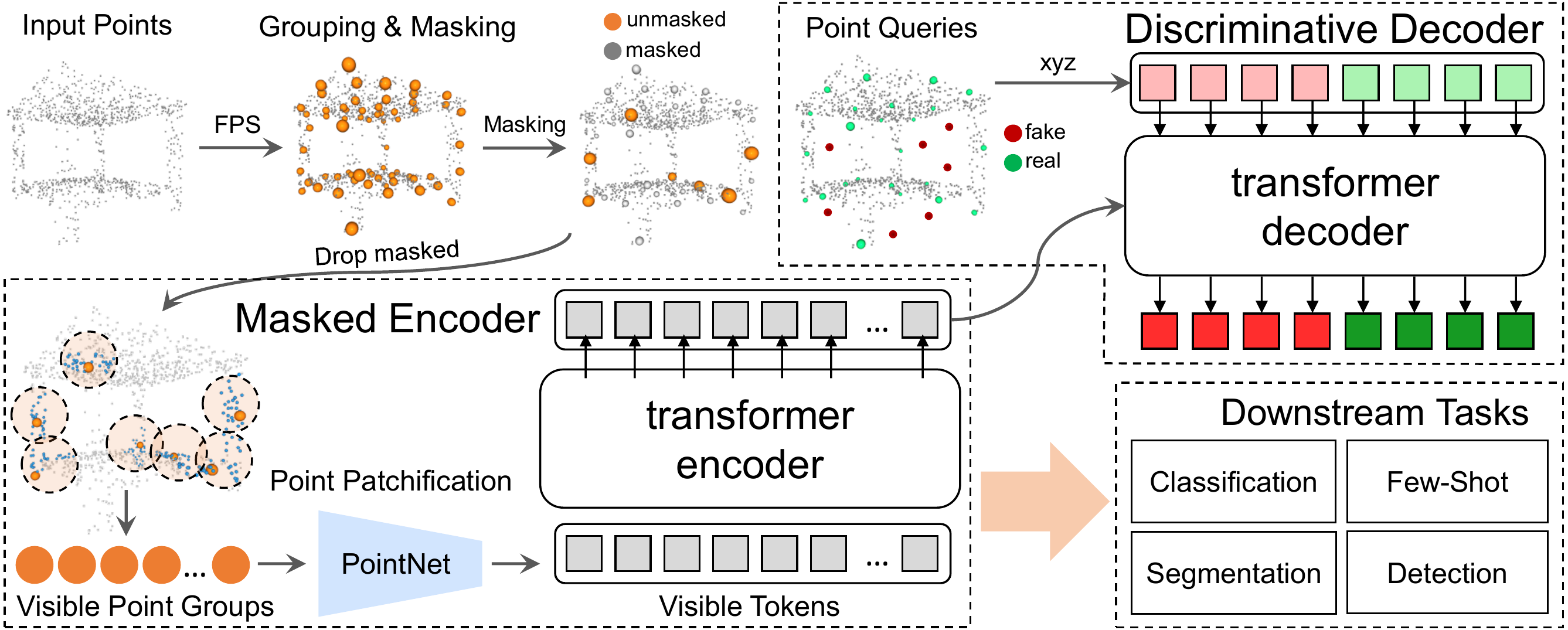}
    \caption{\textbf{\methodname{} architecture.} We first uniformly sample point groups from the point cloud, and partition them to masked and unmasked.  We patchify the visible point groups to token embeddings with PointNet and feed these visible tokens into the encoder.  Then, a set of real query points are sampled from the masked points, and a set of fake query points are randomly sampled from 3D space.  We train the decoder so that it distinguishes between the real and fake points. After pre-training, we discard the decoder and use the encoder for downstream tasks. See Sec.~\ref{sec:dmpm} for details.}
    \label{fig:architecture}
\end{figure}

\vspace{-10pt}
\subsubsection{Transformer Architecture.}
Our network architecture is shown in Fig.~\ref{fig:architecture}. We adopt the standard Transformer encoder~\cite{NIPS2017_3f5ee243} as the encoding  backbone, where each Transformer encoder block consists of a multi-head self-attention (MSA) layer and a feed forward network (FFN). 
As noted earlier, we construct patch-wise features $\{\mathbf{f}_i\}_{i=1}^M$ from the input point cloud $\mathcal{P} \in \mathbb{R}^{N \times 3}$.  Following~\cite{yu2021point}, we apply the MLP positional embedding $\{\mathbf{pos}_i\}_{i=1}^M$  to  the patch features $\{\mathbf{f}_i\}_{i=1}^M$. Then, the class token $\mathbf{E}[s]$, which will be used for downstream classification tasks, is stacked to the top of the patch features $\{\mathbf{f}_i\}_{i=1}^M$;
i.e., the input to the Transformer encoder is $I_0=\left\{\mathbf{E}[\mathrm{s}], \mathbf{f}_{1} +  \mathbf{pos}_1, \cdots, \mathbf{f}_{M} +\mathbf{pos}_M \right\}$. After $n$ Transformer blocks, we get the feature embedding for each point patch $I_n=\left\{\mathbf{E}^n[\mathrm{s}], \mathbf{f}^n_{1} , \cdots, \mathbf{f}^n_{M}  \right\}$. 

During the decoding stage, $N_q$ \textit{real} query points  $\mathbf{Q}_{real}$  and $N_q$ \textit{fake} query points $\mathbf{Q}_{fake}$ are sampled. 
We pass the encoder output $I_n$ and its positional embedding $\{\mathbf{pos}_i\}_{i=1}^M$ , $\mathbf{Q}_{real}$, $\mathbf{Q}_{fake}$ and their positional embedding $\{\mathbf{pos}^{\mathbf{Q}}_i\}_{i=1}^{2N}$ into a one-layer Transformer decoder.
Cross attention is only performed between the queries and encoder keys/values, but not between different queries.
Finally, the decoder output goes through an MLP classification head, which is trained with the binary focal loss~\cite{focalloss}, since there can be a large imbalance between positive and negative samples.

For downstream tasks, the point patchification module and Transformer encoder will be used with their pretrained weights as initialization. 

\vspace{-10pt}
\subsubsection{An Information Theoretic Perspective.}
Here, we provide an information theoretic perspective to our self-supervised learning objective, using mutual information. The mutual information between random variables $X$ and $Y$, $I(X; Y)$, measures the amount of information that can be gained about random variable $X$ from the knowledge about the other random variable $Y$. 

Ideally, we would like the model to learn a rich feature representation of the point cloud: the latent representation $\mathcal{L}$ from our encoder $\mathcal{E}$ should contain enough information to recover the original point cloud $\mathcal{P}$, \textit{i.e.}, we would like to maximize the mutual information $I(\mathcal{P}; \mathcal{L})$. However, directly estimating $I(\mathcal{P}; \mathcal{L})$ is hard since we need to know the exact probability distribution of $P(\mathcal{P} | \mathcal{L})$.  Following \cite{chen2016infogan}, we instead use auxiliary distribution $Q$ to approximate it:
\begin{equation}
\begin{aligned}
I(\mathcal{P}; \mathcal{L}) &=-H(\mathcal{P} | \mathcal{L}) + H(\mathcal{P}) \\
&=\mathbb{E}_{x \sim \mathcal{L}}[\mathbb{E}_{p^{\prime} \sim P(\mathcal{P} | \mathcal{L})}[\log P(p^{\prime} | x)]]+H(\mathcal{P}) \\
&=\mathbb{E}_{x \sim \mathcal{L}}[\underbrace{D_{\mathrm{KL}}(P(\cdot | x) \| Q(\cdot | x))}_{\geq 0} + \mathbb{E}_{p^{\prime} \sim P(\mathcal{P}|\mathcal{L})}[\log Q(p^{\prime} | x)]]+H(\mathcal{P}) \\
& \geq \mathbb{E}_{x \sim \mathcal{L}}[\mathbb{E}_{p^{\prime} \sim P(\mathcal{P}|\mathcal{L})}[\log Q(p^{\prime} | x)]]+H(\mathcal{P})
\end{aligned}
\end{equation}
\textbf{Lemma 3.1.} For random variables $X, Y$ and function $f(x, y)$ under suitable regularity conditions: $\mathbb{E}_{x \sim X, y \sim Y \mid x}[f(x, y)]=\mathbb{E}_{x \sim X, y \sim Y|x, x^{\prime} \sim X| y}[f(x^{\prime}, y)]$.

Therefore, we can define a variational lower bound, $L_{I}(Q, \mathcal{L})$, of the mutual information, $I(\mathcal{P}; \mathcal{L})$:
\begin{equation}
    \begin{aligned}
L_{I}(Q, \mathcal{L}) &=\mathbb{E}_{p \sim P(\mathcal{P}), x \sim \mathcal{L}}[\log Q(p|x)]
+ H(\mathcal{P}) \\
&=\mathbb{E}_{x \sim \mathcal{L}}[\mathbb{E}_{p^{\prime} \sim P(\mathcal{P}|\mathcal{L})}[\log Q(p^{\prime} | x)]] + H(\mathcal{P}) \\
& \leq I(\mathcal{P}; \mathcal{L})
\end{aligned}
\end{equation}
Therefore, we have:
\begin{equation}
\max I(\mathcal{P}; \mathcal{L}) \iff \max L_{I}(Q, \mathcal{L}) \iff \max \mathbb{E}_{x \sim \mathcal{L}}[\mathbb{E}_{p^{\prime} \sim P(\mathcal{P}|\mathcal{L})}[\log Q(p^{\prime} | x)]]
\end{equation}

Previous works use the Chamfer distance to approximate such auxiliary function $Q$, but it has the disadvantage of being sensitive to point sampling variance (discussed in detail in Sec.~\ref{sec:why_not}).  Thus, we instead represent the point cloud distribution with occupancy values within the tightest 3D bounding box of the point cloud:
$\mathcal{B} \in \{x,y,z,o\}^{L}$, where $(x,y,z) \in \mathbb{R}^3$, $o \in \{0, 1\}$, and $L$ is the number of densely sampled points. We let the output of $Q$ denote the continuous distribution of the occupancy value $\hat{o}$, where $Q(\cdot) \in [0, 1]$.
In our implementation, as discussed previously, we construct a set of real query points and fake query points, assign them with the corresponding occupancy labels, and optimize the probability outputs from the model with a binary classification objective.

\subsection{Why not reconstruction, as in MAE?}
\label{sec:why_not}

In this section, we delve into the details on why a reconstruction objective (i.e., reconstructing the original point cloud from the unmasked points) as used in the related Masked AutoEncoder (MAE)~\cite{mae} approach for images would not work for our point cloud setting.  

First, in MAE, the self-supervised learning task is to reconstruct the masked patches, based on the input image's unmasked (visible) patches.  Specifically, given the 2D spatial position for each masked image patch query, the objective is to generate its RGB pixel values. In our case, the analogue would be to generate the spatial xyz values for a masked 3D point patch query -- which would be trivial for the model since the query already contains the corresponding spatial information.  Such a trivial solution will result in perfect zero-loss, and prevent the model from learning meaningful feature representations.

Another issue with the reconstruction objective for point clouds is that there will be point sampling variance. Specifically, the true 3D shape of the object will be a continuous surface, but a point cloud will a discrete sampling of it.  Suppose we sample two such point clouds, and denote the first set as the ``ground truth'' target, and the second set as the prediction of a model.  Although both sets reflect the same geometric shape of the object, the Chamfer distance (which can be used to measure the shape difference between the two point sets) between them is non-zero (i.e., there would be a loss).  Thus, minimizing the Chamfer distance would force the model to generate predictions that exactly match the first set.  And since the first set is just one sampling of the true underlying distribution, this can be an unnecessarily difficult optimization problem that can lead to suboptimal model performance. 

\section{Experiments}

We evaluate the pre-trained representation learned by the proposed model on a variety of downstream point cloud understanding tasks, including object classification, part segmentation, object detection, and few-shot object classification.  We also visualize the reconstruction results from masked point clouds, to qualitatively study the effect of our pretraining.  Finally, we perform ablation studies on masking strategies and decoder designs.

\vspace{-10pt}
\subsubsection{Pretraining Datasets.} \label{sec:datasets}
(1) ShapeNet~\cite{chang2015shapenet} has 50,000 unique 3D models from 55 common object categories, and is used as our pre-training dataset for object classification, part segmentation, and few-shot classification.  For ShapeNet pretraining, we sample 1024 points from each 3D model as the inputs.  We follow \cite{yu2021point} to sample 64 point groups, each containing 32 points.

(2) We also use single-view depth map videos from the popular ScanNet~\cite{dai2017scannet} dataset, which contains around 2.5 million RGBD scans.  We do not use its RGB information in this paper.
We adopt similar pre-processing steps as DepthContrast~\cite{Zhang_2021_ICCV}, but we generate a smaller subset of the dataset than in \cite{Zhang_2021_ICCV}, which we call `ScanNet-Medium', to accelerate pretraining.  ScanNet-Medium is generated by sampling every 10-th frame from ScanNet, resulting in $\sim$25k samples. We use ScanNet-Medium (only geometry information) as the pre-training dataset for 3D object detection.  For pretraining, we sample 20k points from each 3D scene scan as the input.  We follow \cite{misra2021-3detr} to sample 2048 groups, each containing 64 points.

\vspace{-10pt}
\subsubsection{Transformer Encoder.}  We construct a 12-layer standard Transformer encoder, named PointViT, for point cloud understanding.  Following Point-BERT \cite{yu2021point}, we set the hidden dimension of each encoder block to 384, number of heads to 6, FFN expansion ratio to 4, and drop rate of stochastic depth~\cite{huang2016deep} to 0.1.

\vspace{-10pt}
\subsubsection{Transformer Decoder.}  We use a single-layer Transformer decoder for  pretraining.  The configuration of the attention block is identical to the encoder.

\vspace{-10pt}
\subsubsection{Training Details.} Following \cite{yu2021point}, we pretrain with the AdamW~\cite{loshchilov2017decoupled} optimizer with a weight decay of 0.05 and a learning rate of $5\times10^{-4}$ with the cosine decay. The model is trained for 300 epochs with a batch size of 128, with random scaling and translation data augmentation.  Following \cite{yu2021point}, MoCo loss~\cite{he2020momentum} is used for ShapeNet pretraining. For finetuning and other training details, please see supp.

\subsection{3D Object Classification}
\subsubsection{Datasets.}
We compare the performance of object classification on two datasets: the synthetic ModelNet40 \cite{wu20153d}, and real-world ScanObjectNN \cite{uy2019revisiting}.
ModelNet40 \cite{wu20153d} consists of 12,311 CAD models from 40 classes. We follow the official data splitting scheme in \cite{wu20153d}. We evaluate the overall accuracy (OA)  over all test samples.
ScanObjectNN~\cite{uy2019revisiting} is a more challenging point cloud benchmark that consists of 2902 unique objects in 15 categories collected from noisy real-world scans.  It has three splits: OBJ (object only), BG (with background), PB (with background and manually added perturbations). We evaluate the overall accuracy (OA) over all test samples on all three splits.

\begin{table}[t]
\parbox{.51\linewidth}{
    \centering
    \begin{tabular}{lccc}
        \hline
        Method & SSL & \#point & OA \\
        \hline
        PointNet~\cite{qi2017pointnet} & & 1k & 89.2 \\
        PointNet++~\cite{qi2017pointnetplusplus} & & 1k & 90.7 \\
        PointCNN~\cite{li2018pointcnn} & & 1k & 92.2 \\
        SpiderCNN~\cite{xu2018spidercnn} & & 1k & 92.4 \\
        PointWeb~\cite{zhao2019pointweb} & & 1k & 92.3 \\
        PointConv~\cite{wu2019pointconv} & & 1k & 92.5 \\
        DGCNN~\cite{wang2019dynamic} & & 1k & 92.9 \\
        KPConv~\cite{thomas2019kpconv} & & 1k & 92.9 \\
        DensePoint~\cite{liu2019densepoint} & & 1k & 93.2 \\
        PosPool~\cite{liu2020closer} & & 5k & 93.2 \\
        RSCNN~\cite{liu2019relation} & & 5k & 93.6 \\
        \hline
        $[$T$]$ Point Trans.~\cite{engel2021point} & & 1k & 92.8 \\
        $[$T$]$ Point Trans.~\cite{zhao2021point} & & -- & 93.7 \\
        $[$T$]$ PCT~\cite{guo2021pct} & & 1k & 93.2 \\
        $[$ST$]$ PointViT & & 1k & 91.4 \\
        $[$ST$]$ PointViT-OcCo~\cite{occo} & \checkmark & 1k & 92.1 \\
        $[$ST$]$ Point-BERT~\cite{yu2021point} & \checkmark & 1k & 93.2 \\
        $[$ST$]$ \methodname~(Ours) & \checkmark & 1k & \textbf{93.8} \\
        \hline
    \end{tabular}
    \caption{\textbf{Shape Classification on ModelNet40~\cite{wu20153d}.}  With a standard Transformer backbone, our approach significantly outperforms training-from-scratch baselines and SOTA pretraining methods.  It even outperforms PointTransformer~\cite{zhao2021point}, which uses an attention operator specifically designed for point clouds. *SSL: Self-supervised pretraining.
    $[$T$]$: Transformer-based networks with special designs for point clouds. $[$ST$]$: Standard Transformer network.
    }
    \label{tab:modelnet}
}
\parbox{.02\linewidth}{~}
\parbox{.44\linewidth}{
    \centering
    \begin{tabular}{l@{\hskip 0.15cm}c@{\hskip 0.15cm}c@{\hskip 0.15cm}c}
        \hline
        \multirow{2}{*}{Method} & \multicolumn{3}{c}{OA} \\
         & OBJ & BG & PB \\
        \hline
        PointNet~\cite{qi2017pointnet} & 79.2 & 73.3 & 68.0 \\
        PointNet++~\cite{qi2017pointnetplusplus} & 84.3 & 82.3 & 77.9 \\
        PointCNN~\cite{li2018pointcnn} & 85.5 & 86.1 & 78.5 \\
        SpiderCNN~\cite{xu2018spidercnn} & 79.5 & 77.1 & 73.7 \\
        DGCNN~\cite{wang2019dynamic} & 86.2 & 82.8 & 78.1 \\
        BGA-DGCNN~\cite{uy2019revisiting} & -- & -- & 79.7 \\
        BGA-PN++~\cite{uy2019revisiting} & -- & -- & 80.2 \\
        \hline
        PointViT & 80.6 & 79.9 & 77.2 \\
        PointViT-OcCo~\cite{occo} & 85.5 & 84.9 & 78.8 \\
        Point-BERT~\cite{yu2021point} & 88.1 & 87.4 & 83.1 \\
        \methodname~(Ours) & \textbf{89.7} & \textbf{89.3} & \textbf{84.6} \\
        \hline
    \end{tabular}
    \caption{\textbf{Shape Classification on ScanObjectNN~\cite{uy2019revisiting}.} OBJ: object-only; BG: with background; PB: BG with manual perturbation.}
    \label{tab:scanobject}

    \centering
    \begin{tabular}{l@{\hskip 0.5cm}c@{\hskip 0.5cm}c}
        \hline
        \multirow{2}{*}{Method} & \multicolumn{2}{c}{mIoU} \\
         & cat. & ins. \\
        \hline
        PointNet~\cite{qi2017pointnet} & 80.4 & 83.7 \\
        PointNet++~\cite{qi2017pointnetplusplus} & 81.9 & 85.1 \\
        DGCNN~\cite{wang2019dynamic} & 82.3 & 85.2 \\
        \hline
        PointViT & 83.4 & 85.1 \\
        PointViT-OcCo~\cite{occo} & 83.4 & 85.1 \\
        Point-BERT~\cite{yu2021point} & 84.1 & 85.6 \\
        \methodname~(Ours) & \textbf{84.4} & \textbf{86.0} \\
        \hline
    \end{tabular}
    \caption{\textbf{Part Segmentation on ShapeNetPart~\cite{yi2016scalable}.}  Our  method also works well on dense prediction tasks like segmentation.}
    \label{tab:shapenetpart}
}
\end{table}

\vspace{-10pt}
\subsubsection{ModelNet40 Results.}
Table~\ref{tab:modelnet} shows ModelNet40~\cite{wu20153d} results.  With 1k points, our approach achieves a significant 2.4\% OA improvement compared to training from scratch (PointViT).  It also brings a 1.7\% gain over OcCo~\cite{occo} pretraining, and 0.6\% gain over Point-BERT~\cite{yu2021point} pretraining.  The significant improvement over the baselines indicates the effectiveness of our pre-training method.  Notably, for the first time, with 1k points, a standard vision transformer architecture produces competitive performance compared to sophisticatedly designed attention operators from PointTransformer~\cite{zhao2021point} (93.8\% vs 93.7\%).

\vspace{-10pt}
\subsubsection{ScanObjectNN Results.}
We next conduct experiments using the real-world scan dataset ScanObjectNN~\cite{uy2019revisiting}. Table~\ref{tab:scanobject} shows the results.  Our approach achieves SOTA performance on all three splits.  On the hardest PB split, our approach achieves a large 7.4\% OA improvement compared to training from scratch (Point-ViT).  It achieves a 5.8\% gain over OcCo~\cite{occo} pretraining, and a 1.5\% gain over Point-BERT~\cite{yu2021point} pretraining.  The large improvement over the baselines highlights the transferability of our model's self-supervised representation, as there is a significant domain gap between the clean synthetic ShapeNet~\cite{chang2015shapenet} dataset used for pretraining and the noisy real-world ScanObjectNN~\cite{uy2019revisiting} dataset.

We believe the performance gain over OcCo~\cite{occo} and Point-BERT~\cite{yu2021point} is mainly because of our discriminative pretext task.  OcCo suffers from the sampling variance issue (Sec. \ref{sec:why_not}) as it uses a reconstruction-based objective in pretraining.  Compared to Point-BERT, we do not use the point patch mixing technique, which mixes two different point clouds. This could introduce unnecessary noise and domain shifts to pretraining and harm downstream performance.

\subsection{3D Part Segmentation}
\subsubsection{Dataset.}
ShapeNetPart~\cite{yi2016scalable} consists of 16,880 models from 16 shape categories and 50 different part categories, with 14,006 models for training and 2,874 for testing.
We use the sampled point sets produced by \cite{qi2017pointnetplusplus} for a fair comparison with prior work.  We report per-category mean IoU (cat. mIoU) and mean IoU averaged over all test instances (ins. mIoU).

\vspace{-10pt}
\subsubsection{Results.}
Table~\ref{tab:shapenetpart} shows the results (per-category IoU is in supp). Our approach outperforms the training from the scratch (PointViT) and OcCo-pretraining baselines by 1.0\%/0.9\% in cat./ins. mIoU.  It also produces a 0.3\%/0.4\% gain compared to Point-BERT.  Thanks to our dense discriminative pretraining objective, in which we densely classify points over the 3D space, we are able to obtain good performance when scaling to dense prediction tasks.

\subsection{Few-shot Classification}

\begin{table}[t]
    \centering
    \begin{tabular}{lcccc}
        \hline
        \multirow{2}{*}{Method} & \multicolumn{2}{c}{\textbf{5-way}} & \multicolumn{2}{c}{\textbf{10-way}} \\
        & 10-shot & 20-shot & 10-shot & 20-shot \\
        \hline
        DGCNN~\cite{occo} & $91.8 \pm 3.7$ & $93.4 \pm 3.2$ & $86.3 \pm 6.2$ & $90.9 \pm 5.1$ \\
        DGCNN-OcCo~\cite{occo} & $91.9 \pm 3.3$ & $93.9 \pm 3.2$ & $86.4 \pm 5.4$ & $91.3 \pm 4.6 $ \\
        \hline
        PointViT & $87.8 \pm 5.3$ & $93.3 \pm 4.3$ & $84.6 \pm 5.5$ & $89.4 \pm 6.3 $ \\
        PointViT-OcCo~\cite{occo} & $94.0 \pm 3.6$ & $95.9 \pm 2.3$ & $89.4 \pm 5.1$ & $92.4 \pm 4.6 $ \\
        Point-BERT~\cite{yu2021point} & $94.6 \pm 3.1$ & $96.3 \pm 2.7$ & $91.0 \pm 5.4$ & $92.7 \pm 5.1 $ \\
        \methodname~(Ours) & $\mathbf{95.0 \pm 3.7}$ & $\mathbf{97.2 \pm 1.7}$ & $ \mathbf{91.4 \pm 4.0}$ & $\mathbf{93.4 \pm 3.5}$ \\
        \hline
    \end{tabular}
    \caption{\textbf{Few-shot classfication on ModelNet40~\cite{wu20153d}.}}
    \label{tab:fewshot}
\end{table}

We conduct few-shot classification experiments on ModelNet40~\cite{wu20153d}, following the settings in \cite{yu2021point}.  The standard experiment setting is the ``$K$-way $N$-shot'' configuration, where $K$ classes are first randomly selected, and then $N + 20$ objects are sampled for each class.  We train the model on $K \times N$ samples (support set), and evaluate on the remaining $K \times 20$ samples (query set). We compare our approach with OcCo and Point-BERT, which are the current state-of-the-art.

We perform experiments with 4 settings, where for each setting, we run the train/evaluation on 10 different sampled splits, and report the mean and std over the 10 runs.  Table~\ref{tab:fewshot} shows the results.  Our approach achieves the best performance for all settings.  It demonstrate an absolute gain of 7.2\%/3.8\%/4.6\%/2.5\% over the PointViT training from the scratch baseline.  When comparing to pretraining baselines, it outperforms OcCo by 1.0\%/1.3\%/1.5\%/1.0\%, and outperforms Point-BERT by 0.4\%/0.9\%/0.4\%/0.7\%.  It also clearly outperforms the DGCNN baselines.  Our state-of-the-art performance on few-shot classification further demonstrates the effectiveness of our pretraining approach.

\begin{table}[t]
\tabcolsep=0.2cm
    \centering
\begin{tabular}{lcccc} \hline
Methods & SSL & Pretrained Input & $\mathbf{AP}_{25}$ & $\mathbf{AP}_{50}$ \\ \hline 
VoteNet~\cite{qi2019deep} & & - & $58.6$  & $33.5$ \\
STRL~\cite{huang2021spatio}   & \checkmark & Geo & 59.5 & 38.4  \\ 
Implicit Autoencoder~\cite{yan2022implicit}   & \checkmark & Geo  & 61.5 &  39.8  \\
RandomRooms~\cite{rao2021randomrooms}  & \checkmark & Geo  & 61.3 & 36.2 \\
PointContrast~\cite{xie2020pointcontrast}  & \checkmark & Geo & 59.2 & 38.0 \\
DepthContrast~\cite{Zhang_2021_ICCV} & \checkmark & Geo & 61.3 & -- \\
DepthContrast~\cite{Zhang_2021_ICCV}  & \checkmark & Geo + RGB &  \textbf{64.0} & \textbf{42.9} \\
\hline
3DETR~\cite{misra2021-3detr} & & - & 62.1  &  37.9 \\
Point-BERT~\cite{yu2021point}  & \checkmark & Geo & 61.0 & 38.3 \\
\methodname~(Ours)   & \checkmark & Geo & 63.4 & 40.6 \\
\methodname~(Ours, 12 Enc) & \checkmark & Geo & \textbf{64.2} & \textbf{42.1}  \\

\hline
\end{tabular}
\caption{3D object detection results on ScanNet validation set. The backbone of our pretraining model and Point-BERT~\cite{yu2021point} is 3DETR~\cite{misra2021-3detr}. All other methods use VoteNet~\cite{qi2017pointnet} as the finetuning backbone. Only geometry information is fed into the downstream task. ``Input'' column denotes input type for the pretraining stage. ``Geo'' denotes  geometry information. Note that DepthContrast (Geo + RGB) model uses a heavier backbone (PointNet 3x) for downstream tasks.
}
\label{tab:scannet}
\end{table}

\subsection{3D Object Detection}

Our most closely related work, Point-BERT~\cite{yu2021point} showed experiments only on object-level classification and segmentation tasks.
In this paper, we evaluate a model's pretrained representation on a more challenging scene-level downstream task: 3D object detection on  ScanNetV2~\cite{dai2017scannet}, which consists of real-world richly-annotated 3D reconstructions of indoor scenes. It comprises 1201 training scenes, 312 validation scenes and 100 hidden test scenes. Axis-aligned bounding box labels are provided for 18 object categories. 

For this experiment, we adopt 3DETR~\cite{misra2021-3detr} as the downstream model for both our method and Point-BERT. 3DETR is an end-to-end transformer-based 3D object detection pipeline.
During finetuning, the input point cloud is first downsampled to 2048 points via a VoteNet-style Set Aggregation~(SA) layer~\cite{qi2019deep,qi2017pointnetplusplus}, which then goes through 3-layer self-attention blocks. The decoder is composed of 8-layer cross-attention blocks. For a fair comparison with the 3DETR train-from-scratch baseline, we strictly follow its architecture of SA layer and encoder during pretraining, whose weights are transferred during finetuning.
Our pretraining dataset is ScanNet-Medium, as described in Sec.~\ref{sec:datasets}.

Table~\ref{tab:scannet} shows that our method surpasses the 3DETR train-from-scratch baseline by a large margin (+1.3AP$_{25}$ and +2.7AP$_{50}$). Interestingly, Point-BERT brings nearly no improvement compared to training from scratch. The low mask rate and discrete tokens learned from dVAE may impede Point-BERT from learning meaningful representations for detection.
Also, the 3DETR paper~\cite{misra2021-3detr} finds that increasing the number of encoding layers in 3DETR brings only a small benefit to its detection performance. Here we increase the number of layers from 3 to 12, which leads to a large performance improvement (+2.1AP$_{25}$ and +4.2AP$_{50}$) for our approach compared to training from scratch. This result demonstrates that by pre-training on a large unlabeled dataset, we can afford to increase the model's encoder capacity to learn richer representations. Finally, note that we also include VoteNet based methods at the top of Table~\ref{tab:scannet} as a reference, but the numbers are not directly comparable as they are using a different (non Transformer-based) detector.

\subsection{Reconstruction Quality}

\begin{figure}[t!]
    \centering
    \includegraphics[width=.9\textwidth]{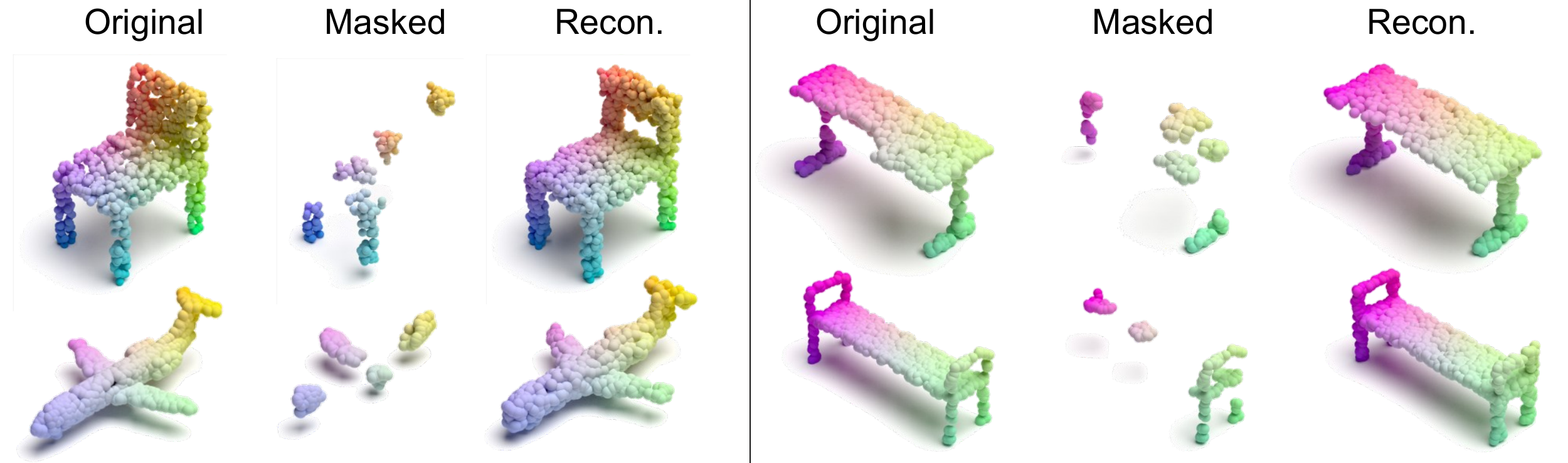}
    \caption{\textbf{Reconstruction results.}
    By reformulating reconstruction as a discriminative occupancy classification task, we achieve a similar learning objective to generative reconstruction while being robust to point sampling variance.
    Even with a high 90\% mask ratio, our approach recovers the overall shape of the original point cloud, without overfitting. Visualization toolkit: Pointflow~\cite{pointflow}.
    }
    \label{fig:reconstruction}
\end{figure}

Although our model is not trained with the reconstruction objective, we can still reconstruct the point cloud with our decoder by classifying the densely sampled points from the point cloud's full 3D bounding box space: $\mathcal{P}_{rec} = \{x|\mathcal{D}(x|\mathcal{L})=1\}$.
Fig.~\ref{fig:reconstruction} shows that even with a large 90\% mask ratio, our model is able to reconstruct the overall shape of the original point cloud, without overfitting.
We also quantitatively evaluate the reconstruction quality with the L2 chamfer distance (CD-$\ell_2$), a standard metric to measure the similarity between two point clouds.  MaskPoint achieves a satisfactory reconstruction accuracy of $6.6\times10^{-3}$ CD-$\ell_2$ (with points $p \in [-1, 1]^3$).
The imperfection is expected (similar to the blurry reconstructed image in MAE~\cite{mae}), as there can be multiple plausible reconstructions given such a large 90\% mask ratio.

\subsection{Ablation Studies}

\begin{table}[t]
    \tabcolsep=0.2cm
    \begin{subtable}[t]{.25\linewidth}
        \centering
        \begin{tabular}[t]{cc}
            ratio & OA \\
            \shline
            0.25 & 83.2 \\
            0.50 & 83.7 \\
            0.75 & 84.1 \\
            0.90 & \textbf{84.6} \\
        \end{tabular}
        \subcaption{\textbf{Random mask}}
        \label{tab:ablation_ratio_random}
    \end{subtable}
    \begin{subtable}[t]{.24\linewidth}
        \centering
        \begin{tabular}[t]{cc}
            ratio & OA \\
            \shline
            0.25 & 82.4 \\
            0.50 & 83.8 \\
            0.75 & 83.7 \\
            0.90 & \textbf{84.1} \\
        \end{tabular}
        \subcaption{\textbf{Block mask}}
        \label{tab:ablation_ratio_block}
    \end{subtable}
    \begin{subtable}[t]{.24\linewidth}
        \centering
        \begin{tabular}[t]{cc}
            \# queries & OA \\
            \shline
            64 & 83.7 \\
            256 & \textbf{84.6} \\
            1024 & 83.9 \\
            \\
        \end{tabular}
        \subcaption{\textbf{\# dec. queries}}
        \label{tab:ablation_dec_queries}
    \end{subtable}
    \begin{subtable}[t]{.24\linewidth}
        \centering
        \begin{tabular}[t]{cc}
            \# dec. & OA \\
            \shline
            1 & \textbf{84.6} \\
            3 & 83.7 \\
            6 & 83.9 \\
            \\
        \end{tabular}
        \subcaption{\textbf{\# dec. layers}}
        \label{tab:ablation_dec_layers}
    \end{subtable}
    \caption{\textbf{Ablations} on ScanObjectNN~\cite{uy2019revisiting} (PB split).  Our findings are: a larger masking ratio generally yields better performance; random masking is slightly better than block masking; 256-query provides a good balance between information and noise; a thin decoder ensures rich feature representation in the encoder and benefits downstream performance.}
\end{table}

\subsubsection{Masking Strategy.}
We show the influence of different masking strategies in Table~\ref{tab:ablation_ratio_random}, \ref{tab:ablation_ratio_block}. First, we observe that a higher masking ratio generally yields better performance, regardless of sampling type.  This matches our intuition that a higher masking ratio creates a harder and more meaningful pretraining task.  Further, with a higher masking ratio, random masking is slightly better than block masking.  Therefore, we use a high mask rate of 90\% with random masking.

\vspace{-10pt}
\subsubsection{Pretraining Decoder Design.}
We study the design of the pretraining decoder in Table~\ref{tab:ablation_dec_queries}, \ref{tab:ablation_dec_layers}, by varying the number of decoder queries and layers. The number of decoder queries influence the balance between the classification of the real points and fake points.  We find that 256-query is the sweet spot, where more queries could introduce too much noise, and fewer queries could result in insufficient training information.

The modeling power of the decoder affects the effectiveness of the pretraining: ideally, we want the encoder to only encode the features, while the decoder only projects the features to the pre-training objective.  Any imbalance in either way can harm the model's performance on downstream tasks.  We find that a single-layer decoder is sufficient for performing our proposed point discrimination task, and having more decoder layers harms the model's performance.

\section{Conclusion}
We proposed a discriminative masked point cloud pretraining framework, which  facilitates a variety of downstream tasks while significantly reducing the pretraining time compared to the prior Transformer-based state-of-the-art method. We adopted occupancy values to represent the point cloud, forming a simpler yet effective binary pretraining objective function. Extensive experiments on 3D shape classification, detection, and segmentation demonstrated the strong performance of our approach. Currently, we randomly mask local point groups to partition the point cloud into masked and unmasked sets. It could be interesting to explore ways to instead learn how to mask the points. We hope our research can raise more attention to mask based self-supervised learning on point clouds. 

\vspace{-10pt}
\subsubsection{Acknowledgement.} This work was supported in part by NSF CAREER IIS-2150012 and the Wisconsin Alumni Research Foundation. We thank Xumin Yu for the helpful discussion in reproducing the Point-BERT baselines.

\appendix

\section*{Appendix}
\section{More results}
\subsubsection{ShapeNetPart}

In Table~\ref{tab:partseg_supp}, we compare the categorical mIoU on ShapeNetPart with other methods.  With a PointViT backbone, we get the highest class mIoU at 84.4\% and the highest instance mIoU at 86.0\%, outperforming previous self-supervised learning approaches (OcCo~\cite{occo} and Point-BERT~\cite{yu2021point}).  It also outperforms standard train-from-scratch point cloud backbones like PointNet++~\cite{qi2017pointnetplusplus} and DGCNN~\cite{wang2019dynamic}. For all categories, our method either has the highest accuracy or is among the best. Thanks to our dense discriminative pretraining objective, in which we densely classify points over the 3D space, we are able to obtain good performance when scaling to dense prediction tasks like part segmentation.

\begin{table}[h]
    \centering
    \begin{adjustbox}{width=\columnwidth,center}
    \begin{tabular}{l|cc|cccccccccccccccc}
\toprule
Methods& cls. & ins. & aero  & bag & cap & car & chair & earp. & guit. & knif. & lamp  & lapt. & mot. & mug & pist. & rock. & skt.  & table \\
\midrule
PointNet~\cite{qi2017pointnet} & 80.4 & 83.7  & 83.4  & 78.7  & 82.5  & 74.9  & 89.6  & 73.0    & 91.5  & 85.9  & 80.8  & 95.3  & 65.2  & 93.0 & 81.2  & 57.9  & 72.8  & 80.6 \\
PN++~\cite{qi2017pointnetplusplus} & 81.9 & 85.1  & 82.4  & 79.0 & 87.7  & 77.3  & 90.8  & 71.8  & 91.0 & 85.9  & 83.7  & 95.3  & 71.6  & 94.1  & 81.3  & 58.7  & 76.4  & 82.6 \\
DGCNN~\cite{wang2019dynamic} & 82.3 & 85.2  & 84.0 & 83.4  & 86.7  & 77.8  & 90.6  & 74.7  & 91.2  & 87.5  & 82.8  & 95.7 & 66.3  & 94.9  & 81.1  & 63.5  & 74.5  & 82.6 \\
\midrule
PointViT & 83.4 & 85.1  & 82.9  & \underline{85.4} & 87.7  & 78.8  & 90.5  & \underline{80.8}  & 91.1  & 87.7 & \underline{85.3}  & \textbf{95.6}  & 73.9  & \underline{94.9} & 83.5  & 61.2  & 74.9  & 80.6 \\
OcCo~\cite{occo} & 83.4 & 85.1  & 83.3  & 85.2  & \textbf{88.3} &  \underline{79.9} & 90.7  & 74.1  & \textbf{91.9}  & 87.6  & 84.7  & 95.4  & 75.5 & 94.4  & 84.1  & 63.1  & 75.7  & 80.8 \\
PN-BERT~\cite{yu2021point} & \underline{84.1} & \underline{85.6} & \textbf{84.3} & 84.8  & 88.0    & 79.8  & \underline{91.0} & \textbf{81.7} & 91.6  & \textbf{87.9}  & 85.2 & \textbf{95.6}  & \underline{75.6}  & 94.7  & \underline{84.3}  & \underline{63.4} & \underline{76.3} & \underline{81.5} \\
MaskPoint (Ours) & \textbf{84.4} & \textbf{86.0} & \underline{84.2} & \textbf{85.6} & \underline{88.1} & \textbf{80.3} & \textbf{91.2} & 79.5 & \textbf{91.9} & \underline{87.8} & \textbf{86.2} & 95.3 & \textbf{76.9} & \textbf{95.0} & \textbf{85.3} & \textbf{64.4} & \textbf{76.9} & \textbf{81.8} \\
\bottomrule
\end{tabular}
\end{adjustbox}
    \caption{Part segmentation results on ShapeNetPart~\cite{yi2016scalable}.
    \textbf{Bold} and \underline{underline} numbers denote best and second best performance, respectively.
    }
    \vspace{-25pt}
    \label{tab:partseg_supp}
\end{table}

\subsubsection{ScanNet} 

Table~\ref{tab:scannet-class} reports per-class average precision
on 18 classes of ScanNetV2 with a 0.25 box IoU
threshold. Relying on purely geometric information,
our method exceeds 3DETR~\cite{misra2021-3detr} in detecting objects like curtain, garbagebin, table, desk, \textit{etc}, where geometry is a
strong cue for recognition.  These results indicate that our mask based discriminative pretraining framework is effective in learning strong geometric representations.
More importantly, our model outperforms 3DETR on classes where it has relatively low AP, \textit{e.g.}, picture, door, curtain, refrigerator, \textit{etc}, which demonstrates the usefulness of pretraining: with the pretrained knowledge relevant to those hard classes, the model is able to make more accurate predictions than the training from the scratch baseline.

\begin{table}[!htp]
\centering
\begin{adjustbox}{width=\columnwidth,center}
\begin{tabular}{l|c|rrrrrrrrrrrrrrrrrr}\toprule
Model & $\mathbf{AP}_{25}$  &cab. &bed &cha. &sofa &tab. &door &win. &boo. &pic. &cou. &desk &cur. &ref. &sho. &toi. &sink &bat. &gar.\\
\midrule
3DETR~\cite{misra2021-3detr} &62.2&50.2 &87.0 &86.0 &87.1 &61.6 &46.6 &40.1 &54.5 &9.1 &62.8 &69.5 &48.4 &50.9 &68.4 &97.9 &67.6 &85.9 &45.8 \\
Ours &63.4&51.8 &82.5 &85.9 &86.8 &69.8 &50.9 &36.9 &47.3 &10.7 &59.6 &76.3 &65.9 &55.6 &66.4 &99.1 &61.5 &83.7 &49.8 \\
Ours (12$\times$) &64.2&49.5 &81.0 &87.2 &86.3 &65.2 &51.3 &42.6 &56.7 &16.2 &56.8 &73.8 &59.6 &56.0 &77.0 &97.8 &66.6 &85.0 &47.7 \\
\bottomrule
\end{tabular}
\end{adjustbox}
\caption{3D object detection scores per category on the ScanNetV2 dataset, evaluated with bbox mIoU 0.25.  Ours (12$\times$): 12 encoder blocks.}
\label{tab:scannet-class}
\end{table}

\begin{figure}[t!]
    \centering
    \includegraphics[width=\textwidth]{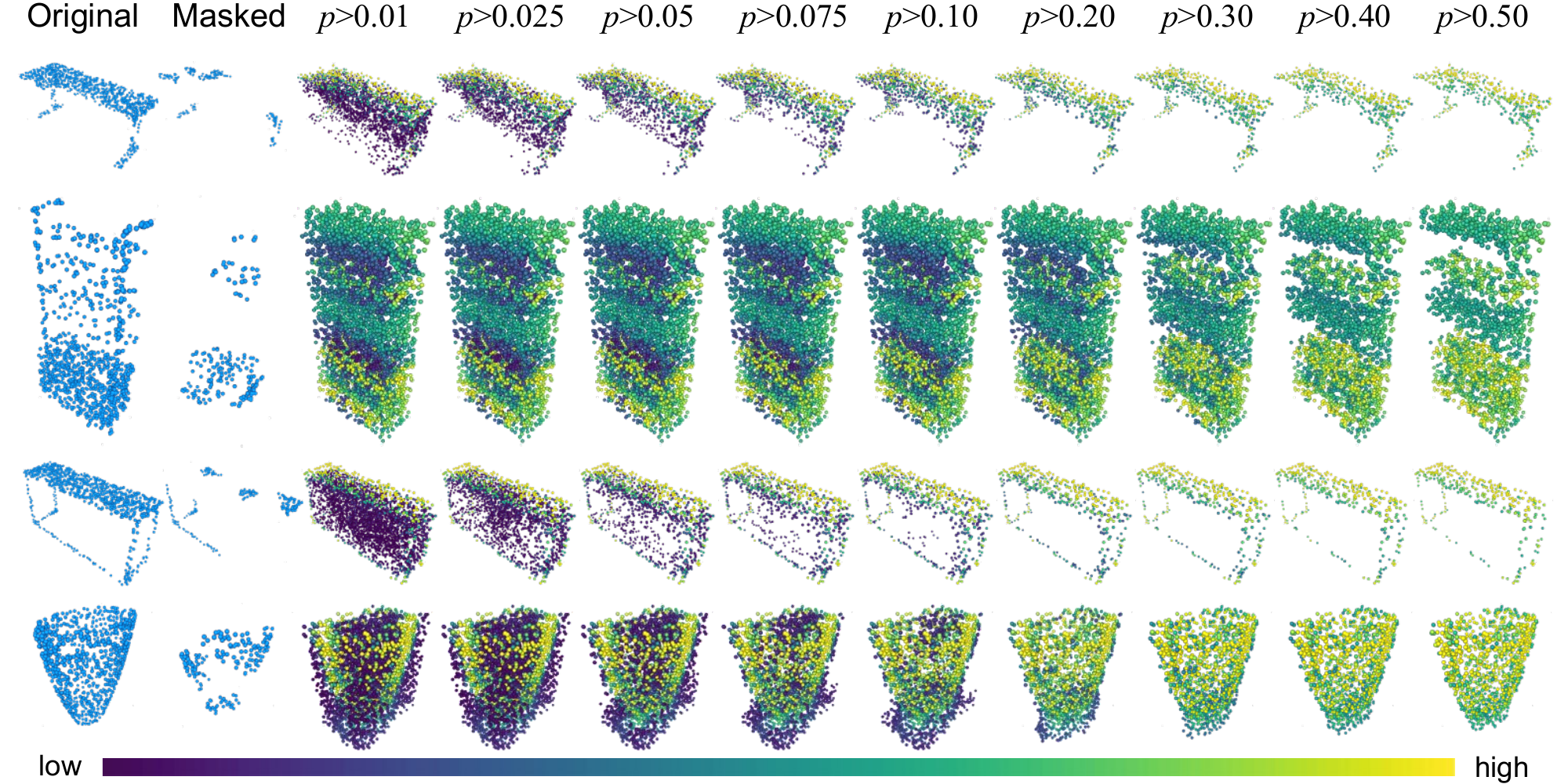}
    \caption{\textbf{Reconstruction results.}
        We densely perform the discriminative occupancy classification task in 3D space, and visualize the predicted occupancy probability.  By varying the confidence threshold $\hat{p}$, we show that our model is able to predict a continuous probability distribution of the occupancy function.
    }
    \label{fig:continuous_reconstruction}
\end{figure}

\subsubsection{More reconstruction visualizations}
We densely perform the discriminative occupancy classification task in 3D space, and visualize in Fig.~\ref{fig:continuous_reconstruction} the predicted occupancy probability.  In different columns, we vary the occupancy threshold $\tau$, and only show the points with occupancy probability prediction that is higher than the given threshold. We can see that our model is able to output a continuous probability distribution of the occupancy function, even if it is only trained with discrete occupancy values from the sampled points.

\begin{figure}[t!]
    \centering
    \includegraphics[width=.8\textwidth]{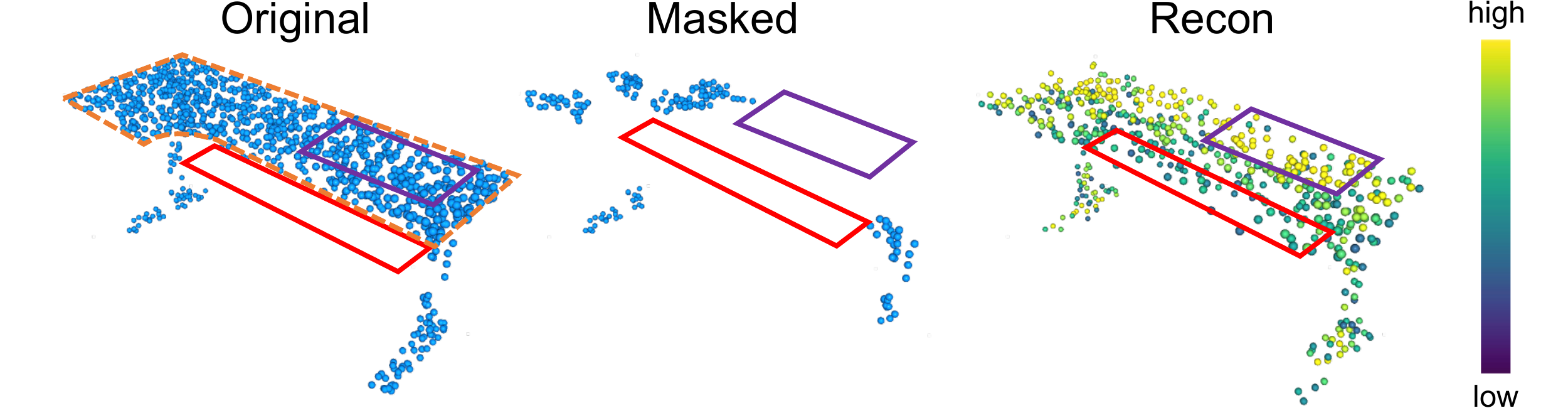}
    \caption{\textbf{A closer look at occupancy distribution.}  Although there are no points present in both {\color{myred} \bf red} and {\color{mypurple} \bf purple} regions of the masked point cloud, the reconstructed probability distribution correctly reflects that of the original point cloud: a lower occupancy in {\color{myred} \bf red} region, and a higher occupancy in {\color{mypurple} \bf purple} region.
    }
    \label{fig:reconstruction_illustrate}
\end{figure}

When we take a closer look at the occupancy distribution, we find several interesting clues on how the model is modeling the probability distribution impressively well. We show our findings in Fig.~\ref{fig:reconstruction_illustrate}.
There are no points present in both {\color{myred} red} and {\color{mypurple} purple} regions of the masked point cloud, while in the original point cloud, there are points present in the {\color{mypurple} purple} region, and no points are in the {\color{myred} red} region.
In the reconstructed probability distribution, the model predicts a low occupancy probability in the {\color{myred} red} region, and a high occupancy in the {\color{mypurple} purple} region.

We find such predictions align with how a human might understand the scene.
First, although there are no points in the purple region of the masked point cloud, given the partial view of the top-left region of the desk top, and the regions where the desk legs are present, it is very likely that there are points present in the purple region (top-right region of the desk top).
As for the red region, the model's prediction can be interpreted as follows: usually  desk tops are rectangle-shaped; however, there do exist desks whose surface shrinks inside the region where the person sits.  Given that there is not a decisive evidence that indicates how this particular desk instance is shaped, the model produces predictions with probability around 0.7, which is lower than other regions that are more certain (yellow points in Fig.~\ref{fig:reconstruction_illustrate} ``Recon'', with $p$$>$0.9).

These two intriguing and encouraging visualizations suggest that our pretrained model is capable of modeling a continuous occupancy probability distribution, and it has learned a deep understanding of the input scene.

\begin{figure}[htbp]
   \vspace{-10pt}
    \begin{subfigure}[t]{0.32\textwidth}
      \centering
      \includegraphics[width=\linewidth]{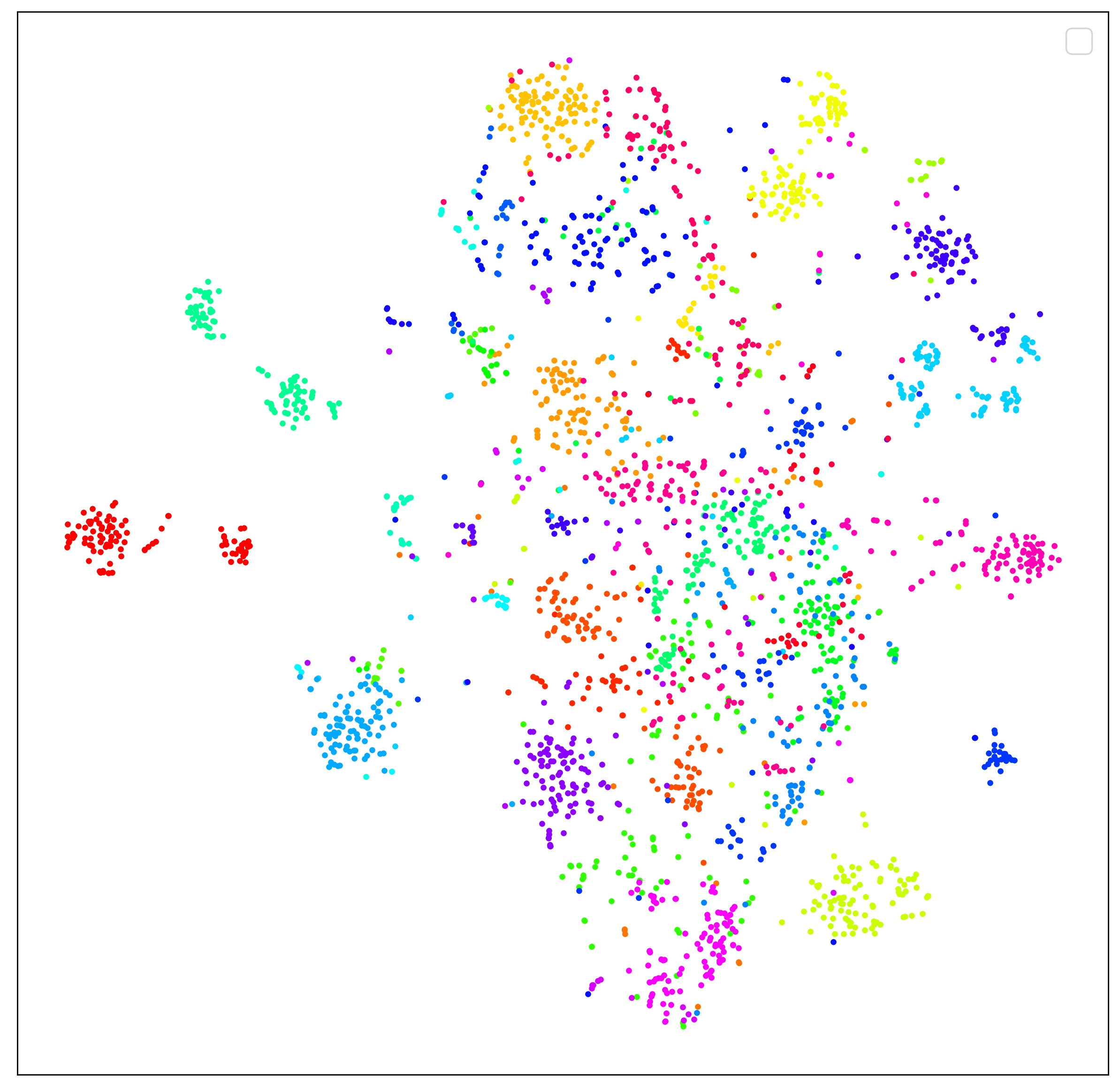}
      \subcaption{Training from scratch}
      \label{fig:sub-second}
    \end{subfigure}
     \begin{subfigure}[t]{0.32\textwidth}
      \centering
\includegraphics[width=\linewidth]{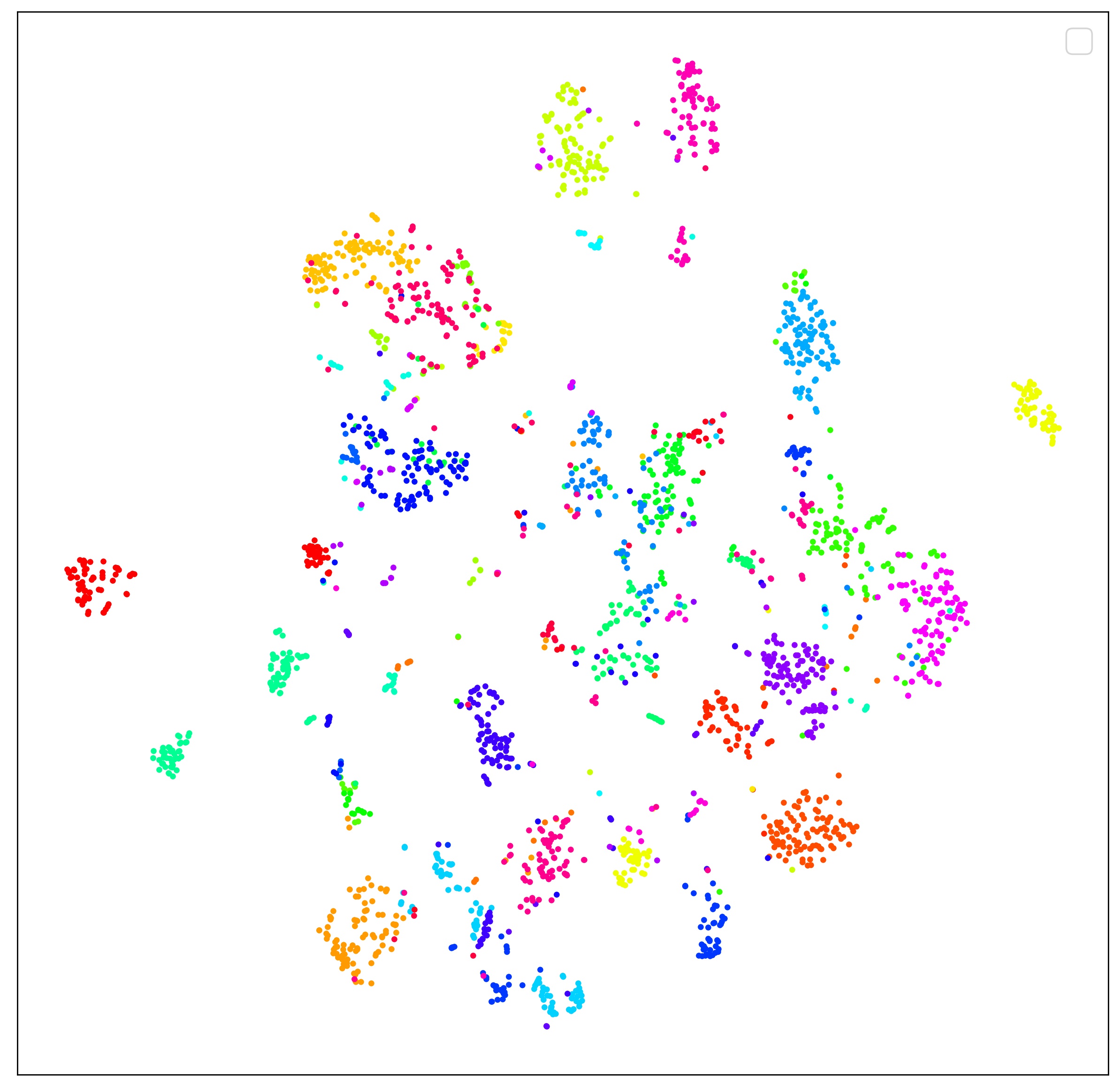}  
      \subcaption{Pretrained on ShapeNet}
      \label{fig:sub-second2}
    \end{subfigure}
    \begin{subfigure}[t]{0.32\textwidth}
      \centering
\includegraphics[width=\linewidth]{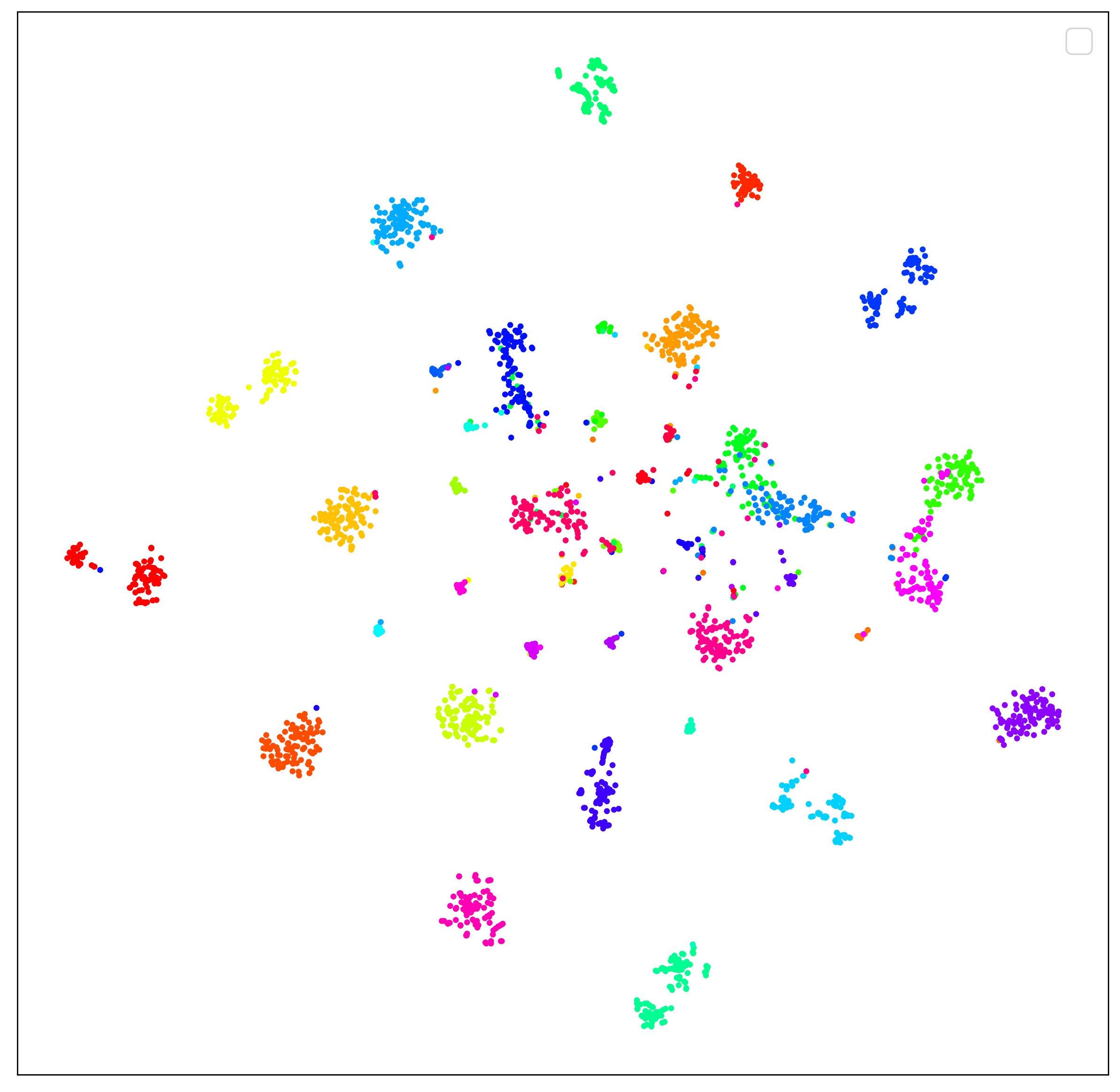}  
      \subcaption{Finetuned on ModelNet}
      \label{fig:sub-second3}
    \end{subfigure}
    \caption{t-SNE visualization of the encoder features for ModelNet40 under three settings: (a) training from scratch, (b) pretrained on ShapeNet, and (c) finetuned on ModelNet40.}
    \vspace{-3ex}
    \label{fig:t_sne}
    \end{figure}
    
\vspace{-15pt}
\subsubsection{t-SNE visualizations}

We show the t-SNE visualizations of the extracted feature vectors from our approach in Fig.~\ref{fig:t_sne}. We use the class token from the encoder output as the high dimensional feature representation for t-SNE. Three setting are adopted here: (a) training from scratch, (b) pretraining on ShapeNet~\cite{chang2015shapenet}, and (c) finetuning on ModelNet40~\cite{wu20153d}.

When training the ModelNet40 classification model from scratch, the resulting features from different categories become heavily entangled, which can leads to less interpretable and robust predictions for new test-time inputs. In contrast, when pretraining the model on ShapeNet using our proposed MaskPoint, the features are much more distinguishable from each other. Furthermore, after finetuning on ModelNet40, the projected features from different classes become clearly separable from each other, which indicates the effectiveness of our approach. Interestingly, the feature clusters in our approach are quite tight. Such feature layout indicates that we can learn a more \textit{compact and disjoint decision boundary}, which has been evaluated to be critical in machine learning applications such mixup~\cite{cutmix,thulasidasan2019mixup} and uncertainty estimation in deep learning~\cite{du2022vos}.

\begin{figure}[t!]
    \centering
    \includegraphics[width=\textwidth]{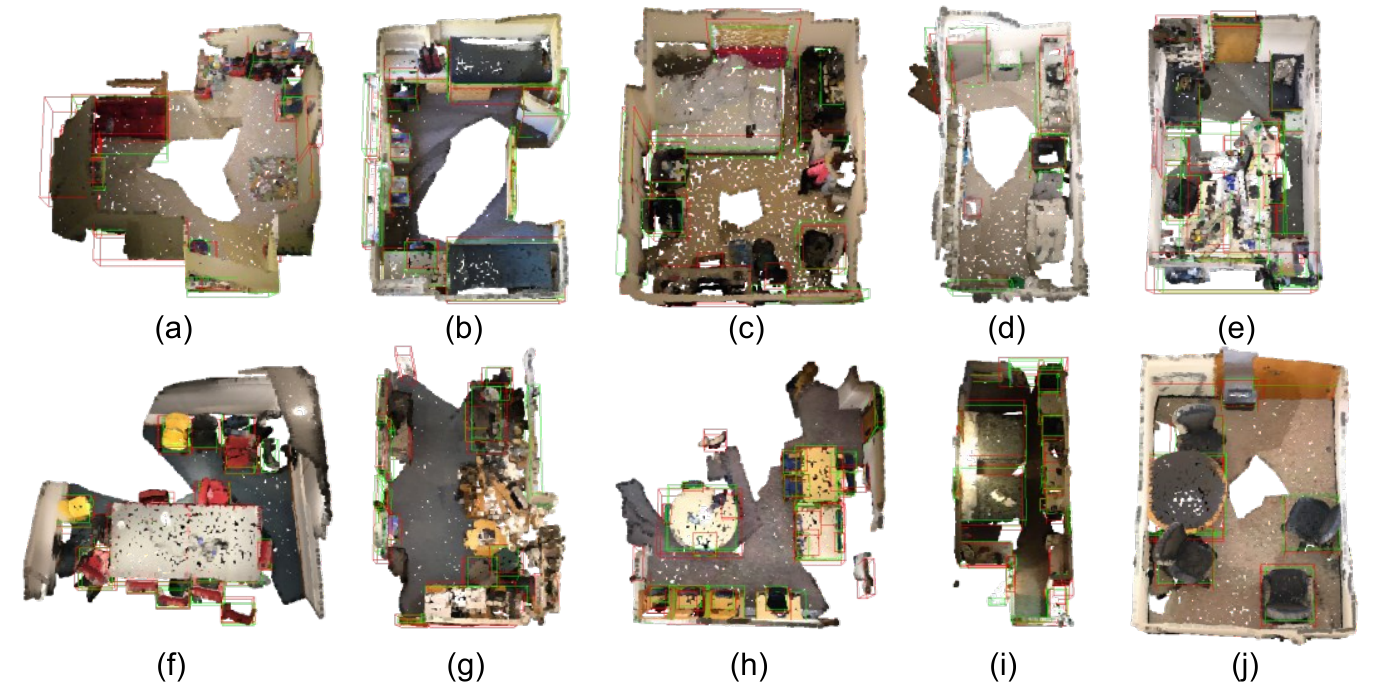}
    \caption{\textbf{Qualitative results of 3D object detection on ScanNetV2~\cite{dai2017scannet}. }
    We show ground truth in {\color{green} \bf green} and predictions in {\color{myred} \bf red} bounding boxes.
    }
    \phantomlabel{a}{fig:scannet_a}
    \phantomlabel{b}{fig:scannet_b}
    \phantomlabel{c}{fig:scannet_c}
    \phantomlabel{d}{fig:scannet_d}
    \phantomlabel{e}{fig:scannet_e}
    \phantomlabel{f}{fig:scannet_f}
    \phantomlabel{g}{fig:scannet_g}
    \phantomlabel{h}{fig:scannet_h}
    \phantomlabel{i}{fig:scannet_i}
    \phantomlabel{j}{fig:scannet_j}
    \label{fig:vis_3detr}
\end{figure}

\vspace{-10pt}
\subsubsection{3D object detection visualizations}
We show 3D object detection visualizations of ScanNetV2 in Figure~\ref{fig:vis_3detr} with {\color{green} green} ground truth bounding boxes and {\color{myred} red} predicted bounding boxes. Our model is capable of precisely localizing the obejct (Fig.~\ref{fig:scannet_b} and Fig.~\ref{fig:scannet_j}).  Results indicate that our masked based discriminative pretraining can not only produce high-quality bounding boxes for the previously annotated objects, but also discover objects that are not annotated. For example, in Fig.~\ref{fig:scannet_c}, our model produces the a bounding box for the bookshelf in the lower region; in Fig.~\ref{fig:scannet_f}, it correctly locates the sofa in the center of the room.
    
\section{Additional Implementation Details}

\subsection{Pretraining}

\subsubsection{Transformer Encoder.}
We follow the standard Transformer design in \cite{NIPS2017_3f5ee243,yu2021point} to construct our point cloud Transformer backbone, PointViT.  It consists of a linear stack of 12 Transformer blocks, where each Transformer block contains a multi-head self-attention (MHSA) layer and a feed-forward network (FFN).  LayerNorm (LN) is adopted in both layers.  Following \cite{yu2021point}, we use the MLP-based positional embedding. We set the Transformer hidden dimension to 384, MHSA head number to 6, expansion rate of FFN to 4, stochastic drop path~\cite{huang2016deep} rate to 0.1.

\vspace{-10pt}
\subsubsection{Feed-forward network (FFN).} Following \cite{yu2021point}, we use a two-layer MLP with ReLU and dropout as the feed-forward network.  Dropout rate is set to 0.1.

\vspace{-10pt}
\subsubsection{Positional Embeddings.}
Following \cite{yu2021point}, we use a two-layer MLP with GELU~\cite{hendrycks2016gaussian} as the positional embedding module.  All Transformer modules share the same positional embedding MLP module.  Detailed configuration is shown in Table~\ref{tab:detail_design_config}.

\vspace{-10pt}
\subsubsection{Point Classification Head.}
We use a simple two-layer MLP with GELU~\cite{hendrycks2016gaussian} for the point classification pretext task in pretraining.  We use the binary focal loss~\cite{focalloss} to balance the information from positive and negative samples. Detailed configuration is shown in Table~\ref{tab:detail_design_config}.

\begin{table}[t]
    \tabcolsep=0.2cm
    \centering
\begin{tabular}{l|c|c|c|c|c|c}
\toprule
Module & Block & $C_{in}$ & $C_{out}$ & $k$ & $N_{out}$ & $C_{middle}$ \\
\midrule
Positional Embed. & MLP & 3 & 384 & & & 128 \\
\midrule
Point Classify Head & MLP & 384 & 2 & & & 64 \\
\midrule
Classification Head & MLP & 768 & $N_{cls}$ & & & 512, 256 \\
\midrule
\multirow{9}{*}{Segmentation Head} & MLP & 387 & 384 & & & \underline{~~~~384$\times$4~~~~}  \\
& DGCNN & 384 & 512 & 4 & 128 & \\
& DGCNN & 512 & 384 & 4 & 128 & \\
& DGCNN & 384 & 512 & 4 & 256 & \\
& DGCNN & 512 & 384 & 4 & 256 & \\
& DGCNN & 384 & 512 & 4 & 512 & \\
& DGCNN & 512 & 384 & 4 & 512 & \\
& DGCNN & 384 & 512 & 4 & 2048 & \\
& DGCNN & 512 & 384 & 4 & 2048 & \\
\bottomrule
\end{tabular}
    \caption{Detailed module design of MaskPoint.  $C_{in}/C_{out}$ denotes the input/output channels, $C_{middle}$ denotes the hidden channels of MLP modules, $N_{out}$ denotes the cardinality of the output point/feature set, $k$ is the number of neighbors used in the $k$-NN operator.}
    \label{tab:detail_design_config}
    \vspace{-8pt}
\end{table}

\vspace{-10pt}
\subsubsection{ScanNet-Medium Pretraining}

Note that for ScanNet-Medium pretraining, we use the encoder with 3 Transformer blocks, where each block still consists of a MHSA layer and a FFN layer. LN and MLP positional embedding are also utilized in the encoder. Following the downstream architecture of 3DETR~\cite{misra2021-3detr}, we set the hidden dimension to be 256, the number of MHSA heads to be 4, and Dropout rate to be 0.1 for the Transformer. The hidden dimension is set to be 128 for the FFN layer. 

For other settings such as positional embedding and classification head, the setting is exactly the same as the ShapeNet pretraining setting. 
    
\subsection{Finetuning}

\subsubsection{Classification}
We use a three-layer MLP with dropout for the classification head. The input feature to the classification head consists of two parts from the Transformer encoder: (1) the CLS token; (2) the max-pooled feature of other output features. These two features are concatenated together and fed into the classification head. For ScanObjectNN dataset experiments, 10-percentile clipping~\cite{seetharaman2020autoclip} is used for stablizing the training.  Detailed configuration is shown in Table~\ref{tab:detail_design_config}.

\vspace{-10pt}
\subsubsection{Part Segmentation}
The standard Transformer only has a single-scale feature output, which is not suitable for common head designs for dense prediction tasks like segmentation.  Following \cite{yu2021point}, after getting the feature outputs from the Transformer encoder, we perform segmentation in two steps: (1) generating a multi-scale feature pyramid from the Transformer encoder outputs; (2) applying a standard feature propagation head for point cloud segmentation on the generated multi-scale feature maps to generate dense predictions.

We obtain the feature maps $f_{\{4,8,12\}} \in \mathbb{R}^{N_3 \times d}$ from the 4th, 8th, 12th layer, and our goal is to convert them to a feature pyramid with different cardinality $N_{\{0,1,2,3\}}$, where $N_0$ is the cardinality of the original point cloud $\mathcal{P}$, and $N_{\{1,2,3\}}$ are the desired cardinality of the feature maps at different scales; in our case, $N_{\{0,1,2,3\}} = \{2048, 512, 256, 128\}$.

First, we use furthest point sampling (FPS) to downsample the original point cloud $\mathcal{P}_0$ to different resolutions: $\mathcal{P}_{\{1,2,3\}} \in \mathbb{R}^{N_{\{1,2,3\}} \times 3}$, then a feature propagation module is used to upsample the feature maps $f_{\{4,8,12\}}$ to the corresponding cardinality  $f^{up}_{\{4,8,12\}} \in \mathbb{R}^{N_{\{1,2,3\}} \times d}$.

After obtaining the multi-scale feature maps, we then apply the DGCNN module to propagate the features through different scales, $\hat{f}_4 = \operatorname{DGCNN}(f^{up}_{\{4,8,12\}})$.  Another feature propagation layer is then applied on $\hat{f}_4$ for upsampling to the highest resolution $\hat{f}_0 \in \mathbb{R}^{N_0 \times d}$.

Finally, we apply a pointwise MLP classifier on the features at the highest resolution $\hat{f}_0$ to obtain the segmentation results.  Detailed configuration is shown in Table~\ref{tab:detail_design_config}.

\begin{table}[t!]
\tabcolsep=0.2cm

\parbox{.45\linewidth}{
    \begin{tabular}{l|l}
        config & value\\
        \shline
        epochs & 300 \\
        optimizer & AdamW \\
        learning rate & 5e-4 \\
        weight decay & 5e-2 \\
        LR schedule & cosine decay \\
        warmup epochs & 3 \\
        augmentation & Scale/Translate \\
        batch size & 128 \\
        \# points & 1024 \\
        \# patches & 64 \\
        patch size & 32 \\
        mask ratio & 0.90 \\
        mask type & random \\
    \end{tabular}
    \vspace{3pt}
    \caption{Pretraining setting on ShapeNet~\cite{chang2015shapenet}.}
    \label{tab:pretrain_shapenet}
}
\hspace{.05\linewidth}
\parbox{.45\linewidth}{
    \centering
    \begin{tabular}[t]{l|c}
        config & value \\
        \shline
        epochs & 300 \\
        optimizer & AdamW \\
        learning rate & 5e-4 \\
        weight decay & 5e-2 \\
        LR schedule & cosine decay \\
        warmup epochs & 10 \\
        augmentation & Scale/Translate \\
        batch size & 32(cls), 16(seg) \\
        \# points & 1024(cls), 2048(seg) \\
        \# patches & 64(cls), 128(seg) \\
        patch size & 32 \\
    \end{tabular}
    \vspace{3pt}
    \caption{Finetuning setting on classification (cls) and segmentation (seg).}
    \label{tab:finetuning}
}
\vspace{-20pt}
\end{table}

\subsubsection{3D Object Detection}
We strictly follow the setting of the original 3DETR~\cite{misra2021-3detr} model as the downstream 3D object detector. The points are first donwsampled to 2048 points using a Set-Aggregation~(SA) layer. The encoder is composed of 3 standard Transformer blocks. The decoder is comprised of 8 Transformer blocks using cross attention. During finetuning, only the weights of the SA layer and the encoder are transferred to the downstream tasks.  The finetuning epoch number is 1080, the optimizer is AdamW with learning rate of $5\times10^{-4}$ and weight decay of 0.1, the batch size is 8.

\bibliographystyle{splncs04}
\bibliography{egbib}

\end{document}